\documentclass{revtex4-2}

\usepackage{graphicx}   
\usepackage{dcolumn}    
\usepackage{bm}         

\usepackage[table,dvipsnames]{xcolor}
\usepackage{booktabs}
\usepackage{siunitx}
\usepackage{hyperref}
\usepackage{amsmath}
\usepackage[capitalize]{cleveref}
\usepackage{caption}
\usepackage{float}
\usepackage{amssymb}
\usepackage{threeparttable}

\usepackage[nolist]{acronym}
\begin{acronym}
    \acro{BO}{Bayesian optimisation}
    \acro{EI}{expected improvement}
    \acro{EA}{Experimental Area}
    \acro{FDF}{focus-defocus-focus}
    \acro{GP}{Gaussian process}
    \acro{MAE}{mean absolute error}
    \acro{MSE}{mean squared error}
    \acro{RMSE}{root mean squared error}
    \acro{MLP}{multilayer perceptron}
    \acro{RL}{Reinforcement Learning}
    \acro{RLO}{Reinforcement Learning-trained Optimisation}
    \acro{TD3}{Twin Delayed DDPG}
    \acro{UCB}{upper confidence bound}
\end{acronym}

\newcommand{\action}{\bm{a}}
\newcommand{\actuators}{\bm{u}}

\newcommand{\kick}{\alpha}
\newcommand{\incomingbeam}{I}
\newcommand{\measuredbeam}{\bm{b}}
\newcommand{\measurementaccuracy}{\epsilon}
\newcommand{\misalignments}{M}

\newcommand{\state}{\bm{s}}

\newcommand{\targetbeam}{\bm{b}'}
\newcommand{\quadstrength}{k}

\graphicspath{ {./figures/} }

\newcommand{\nrealtrials}{\num{22} }

\begin{document}

\title{Learning to Do or Learning While Doing: Reinforcement Learning and Bayesian Optimisation for Online Continuous Tuning}

\author{Jan Kaiser}
\thanks{jan.kaiser@desy.de}
\author{Annika Eichler}
\author{Oliver Stein}
\author{Willi Kuropka}
\author{Hannes Dinter}
\author{Frank Mayet}
\author{Thomas Vinatier}
\author{Florian Burkart}
\author{Holger Schlarb}
\affiliation{Deutsches Elektronen-Synchrotron DESY, Germany}

\author{Chenran Xu}
\thanks{chenran.xu@kit.edu}
\author{Andrea {Santamaria Garcia}}
\author{Erik Bründermann}
\affiliation{Karlsruhe Institute of Technology KIT, Germany}

\date{5 June 2023}

\begin{abstract}
Online tuning of real-world plants is a complex optimisation problem that continues to require manual intervention by experienced human operators. 
Autonomous tuning is a rapidly expanding field of research, where learning-based methods, such as \acf{RLO} and \acf{BO}, hold great promise for achieving outstanding plant performance and reducing tuning times.
Which algorithm to choose in different scenarios, however, remains an open question.
Here we present a comparative study using a routine task in a real particle accelerator as an example, showing that \ac{RLO} generally outperforms \ac{BO}, but is not always the best choice. Based on the study's results, we provide a clear set of criteria to guide the choice of algorithm for a given tuning task. These can ease the adoption of learning-based autonomous tuning solutions to the operation of complex real-world plants,
ultimately improving the availability and pushing the limits of operability of these facilities, thereby enabling scientific and engineering advancements.
\end{abstract}

\maketitle

\acresetall

\section{Introduction}

Complex real-world plants
are instrumental in facilitating scientific and technological progress. For their successful operation, it is critical that these facilities achieve predefined performance metrics. These are reached through online tuning, i.e. the optimisation of the plant and its subsystems towards a desired system state.
Tuning these systems is a challenging optimisation problem due to the non-linear and often dynamic correlations among a large number of tuning parameters.
Moreover, the inherent noise in real-world measurements, the time-consuming data acquisition, and the high costs associated with system downtime make the tuning of real-world systems particularly challenging.

To date, online tuning continues to be performed manually, relying on the experience of expert human operators. This leads to suboptimal solutions that are labour intensive to attain and difficult to reproduce. 

To reduce downtime and push the limits of their operational capabilities, efforts are made to develop autonomous plant tuning solutions. Existing approaches can improve optimisation results, reproducibility, and reliability for some tuning tasks, but come with their own drawbacks.
For example, while grid search and random search are reliable and highly reproducible approaches, they require a large number of samples. As a result, these methods become impractical in the real world, where the cost per sample may be high.
Other approaches from the field of numerical optimisation can reduce the number of required samples and have been successfully applied to tuning tasks~\cite{bergan2019online,huang2013algorithm}.
While these approaches show promising results, their performance drops as the number of tuning dimensions increases due to the so-called \textit{curse of dimensionality}~\cite{bellman1957dynamic}. Furthermore, many of these methods are sensitive to noise~\cite{huang2013algorithm}, which is omnipresent in real-world measurements.

Learning-based methods have emerged as promising solutions capable of sample-efficient, high-dimensional optimisation under real-world conditions. \Ac{BO}~\cite{shahriari2016taking} is one such learning-based method, that has recently risen in popularity.
In \ac{BO}, the number of samples required for a successful optimisation is reduced by learning a surrogate model of the objective function at the time of optimisation. Another promising approach is \textit{optimiser learning}~\cite{andrychowicz2016learning,chen2022learning,li2017learning,li2017learning2}, where the function predicting the next sampling point is learned before the application. A powerful instance of optimiser learning is the use of a neural network that is trained via \ac{RL}~\cite{li2017learning,li2017learning2}, allowing for the automated discovery of optimisation algorithms. In this paper, we call the resulting optimisation algorithm \ac{RLO}. As continued optimisation of a dynamic function can be considered to be equivalent to control, we consider \ac{RLO} to be equivalent to \ac{RL}-based control. 
Both \ac{RLO} and \ac{BO} are very actively researched and applied to a variety of real-world plants such as particle accelerators~\cite{boltz2020feedback,bruchon2020basic,duris2020bayesian,hanuka2019online,jalas2021bayesian,kain2020sampleefficient,kaiser2022learningbased,madysa2022automated,mcintire2016bayesian,oshea2020policy,pang2020autonomous,shalloo2020automation,stjohn2021realtime,xu2023bayesian}, fusion reactors~\cite{degrave2022magnetic,seo2021feedforward,seo2022development}, optical and radio telescopes~\cite{guerraramos2020towards,nousiainen2022toward,yatawatta2021deep}, chemical reactions~\cite{zhou2017optimizing}, additive manufacturing~\cite{deneault2021toward}, photovoltaic power plants~\cite{abdelrahman2016bayesian}, spacecraft~\cite{xiong2020intelligent,xiong2022application}, airborne wind energy systems~\cite{baheri2017realtime}, telecommunication networks~\cite{maggi2021bayesian} and grid-interactive buildings~\cite{ding2020mb2c,nweye2022realworld}, amongst others. 
In each of these fields, both \ac{RLO} and \ac{BO} have achieved excellent tuning results at high sample efficiency.
To the best of our knowledge, however, there is no previous work conducting a detailed comparison of \ac{RLO} and \ac{BO} for online continuous optimisation of real-world plants.

In this work, we study \ac{RLO} and \ac{BO} for tuning a subsystem of a particle accelerator and compare them in terms of the achieved optimisation result and their convergence speed. 
In the field of particle accelerators, both methods are gaining notable attention and have led to significant improvements~\cite{boltz2020feedback,bruchon2020basic,duris2020bayesian,hanuka2019online,jalas2021bayesian,kain2020sampleefficient,kaiser2022learningbased,madysa2022automated,mcintire2016bayesian,oshea2020policy,pang2020autonomous,shalloo2020automation,stjohn2021realtime,xu2023bayesian}. 
To ensure the reliability of our results, we combine a significant number of simulations with real-world measurements.
Based on the results of our study, we ascertain the advantages and disadvantages of each tuning method and identify criteria to guide the choice of algorithm for future applications.

\section{Results}\label{sec:results}
 
\begin{figure}
    \centering
    \includegraphics{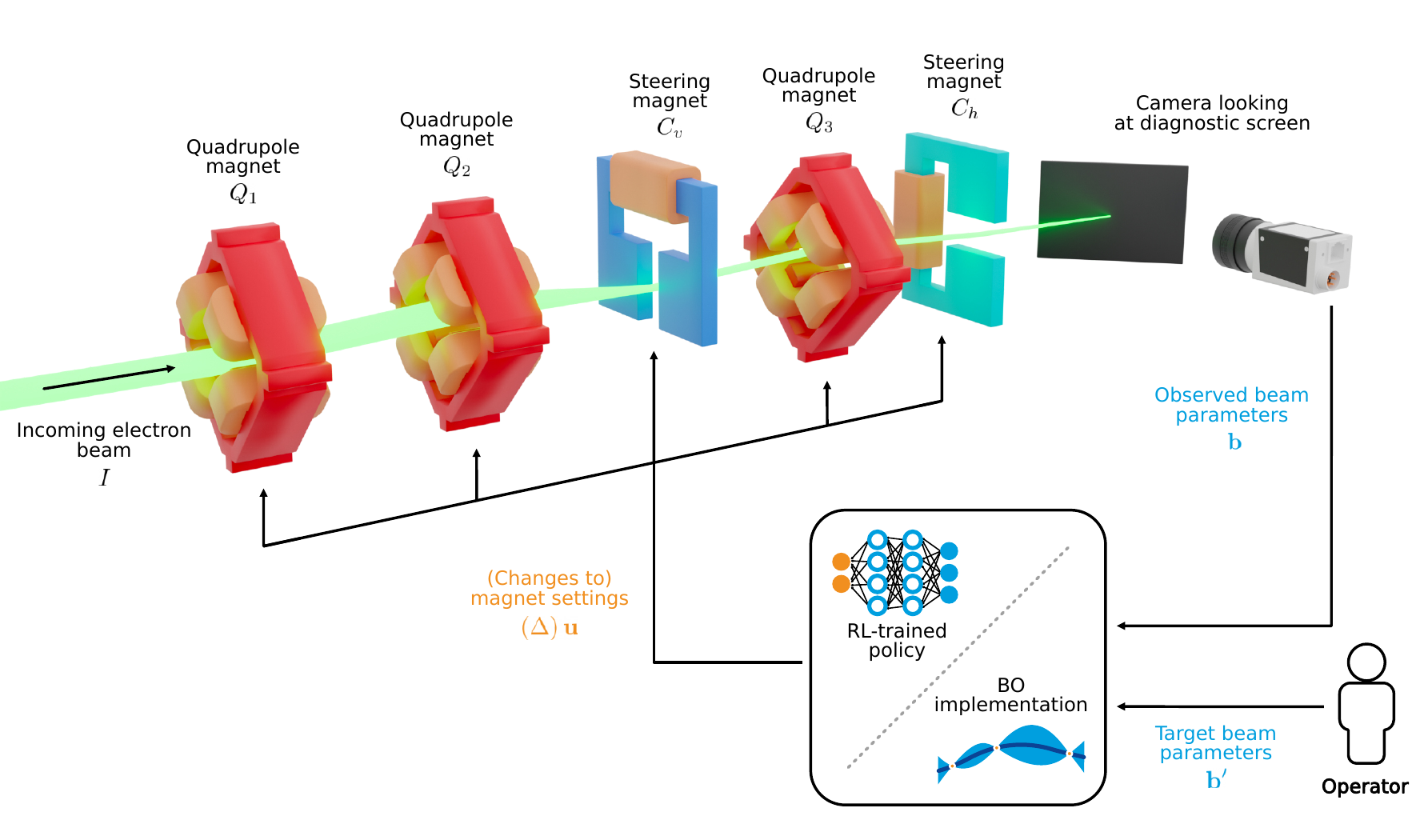}
    \caption{\textbf{Simplified 3D illustration of the considered section of the ARES particle accelerator.} This section consists of three quadrupole magnets and two steering magnets, followed by a diagnostic screen. The measured beam $\measuredbeam$ and the desired beam $\targetbeam$ are provided to the algorithm performing the tuning. In the case of \ac{BO}, they are used to compute the objective. In the case of \ac{RL}, they are provided along with the magnet settings as input to the policy and are used to calculate the reward. Both algorithms output either the next settings to the magnets $\actuators$ or a change to the magnets $\Delta\actuators$.}
    \label{fig:aresea_layout}
\end{figure}

In this study, we consider as a benchmark a recurring beam tuning task which is ubiquitous across linear particle accelerators and frequently performed during start-up and operation mode changes, where the goal is to focus and steer the electron beam on a diagnostic screen. 
While this task can be very time-consuming, it is also well-defined, making it suitable as a proof-of-concept application for \ac{RLO} and \ac{BO}.
For the study, we use the specific magnet lattice of 
a section of
the ARES (Accelerator Research Experiment at SINBAD) particle accelerator~\cite{panofski2021commissioning,burkart2022the} at DESY in Hamburg, Germany, one of the leading accelerator centres worldwide. From here on, we refer to this section as the \textit{accelerator section}. An illustration of the accelerator section is shown in \cref{fig:aresea_layout}.
Further details on ARES and the accelerator section are given in \cref{sec:ares_experimental_area}. 
The lattice of the accelerator section is in downstream order composed of two quadrupole focusing magnets $Q_1$ and $Q_2$, a vertical steering magnet $C_v$, a third quadrupole focusing magnet $Q_3$, and a horizontal steering magnet $C_h$.
Downstream of the magnets there is a diagnostic screen capturing a transverse image of the electron beam. A Gaussian distribution $\measuredbeam = \left( \mu_x, \sigma_x, \mu_y, \sigma_y \right)$ is fitted to the observed image, where $\mu_{x, y}$ denote the transverse beam positions and $\sigma_{x,y}$ denote the transverse beam sizes.
The goal of the tuning task is to adjust the quadrupole magnets' field strengths $k$ and steering magnets' steering angles $\alpha$ to achieve a target beam $\targetbeam$ chosen by a human operator.
We denote the \textit{actuators}, here the magnet settings to be changed by the algorithm, as $\actuators = \left( k_{Q_1}, k_{Q_2}, \alpha_{C_v}, k_{Q_3}, \alpha_{C_h} \right)$.
The optimisation problem can be formalised as minimising the objective

\begin{equation}
    \min O \left( \actuators \right) = \min D \left( \measuredbeam, \targetbeam \right), 
\end{equation}

which for the benchmark tuning task is defined as the difference $D$ between the target beam~$\targetbeam$ and the observed beam~$\measuredbeam$.
The observed beam $\measuredbeam$ is determined by the beam dynamics, which depend on the actuators~$\actuators$, and environmental factors, such as the magnet misalignments and the incoming beam to the accelerator section. Together with the target beam $\targetbeam$, these define the \textit{state} of the environment. With most real-world tuning tasks, not all of the state can be observed, i.e. it is \textit{partially observable}. In the case of the benchmark task, the magnet misalignments and the incoming beam cannot be easily measured or controlled, and are therefore part of the environment's hidden state.
As a measure of difference between the observed beam $\measuredbeam$ and the target beam $\targetbeam$, we use the \ac{MAE} defined as

\begin{equation}\label{eq:mae}
    D_\mathrm{MAE} (\measuredbeam, \targetbeam) = \frac{1}{4} \sum_{i=1}^{4} \left| \measuredbeam^{(i)} - \targetbeam^{(i)} \right|,
\end{equation}

i.e. the mean of the absolute value of the beam parameter differences over all four beam parameters, where $\measuredbeam^{(i)}$ denotes the $i$-th element of $\measuredbeam$.

For this study, an \ac{RLO} policy was trained according to previous work~\cite{kaiser2022learningbased} and as described in \cref{sec:reinforcement_learning}. An implementation of \ac{BO} with a \ac{GP} model~\cite{rasmussen2006gaussian}, detailed in \cref{sec:bayesian_optimisation}, was specially designed for this study.
In addition to the studied \ac{RLO} and \ac{BO} solutions, we consider random search and Nelder-Mead Simplex optimisation~\cite{nelder1965simplex} as baselines for randomised and heuristic optimisation algorithms. They are presented in \cref{sec:random_search,sec:nelder_mead}, respectively.

\subsection{Simulation study}\label{sec:simulation_study}

For the simulation study, we consider a fixed set of \num{300} randomly generated environment states, each defined by a target beam $\targetbeam$, an incoming beam $\incomingbeam$ entering the accelerator section from upstream, and transverse misalignments of the quadrupole magnets and the diagnostic screen $\misalignments$. We refer to these instances of the environment state as \textit{trials}, defined in \cref{eq:trial}.
The results of the simulation study over \ac{RLO} and \ac{BO}, as well as the two baseline algorithms, random search and Nelder-Mead Simplex, are summarised in \cref{tab:results}.

We find that the learning-based algorithms \ac{RLO} and \ac{BO} outperform both baselines in terms of the optimisation result, achieving a final beam difference $D$ 
at least 6 times smaller. 
Furthermore, \ac{RLO} achieves a median final beam difference $D$ of \qty{4}{\micro\meter}, which is more than an order of magnitude smaller than the one achieved by \ac{BO}. The final beam difference achieved by \ac{RLO} is smaller than the one achieved by \ac{BO} in \qty{96}{\percent} of the trials. 
Note that the final beam difference achieved by \ac{RLO} is smaller than the measurement accuracy $\measurementaccuracy = \qty{20}{\micro\meter}$ of the real-world diagnostic screen.

\begin{figure}
    \centering
    \includegraphics{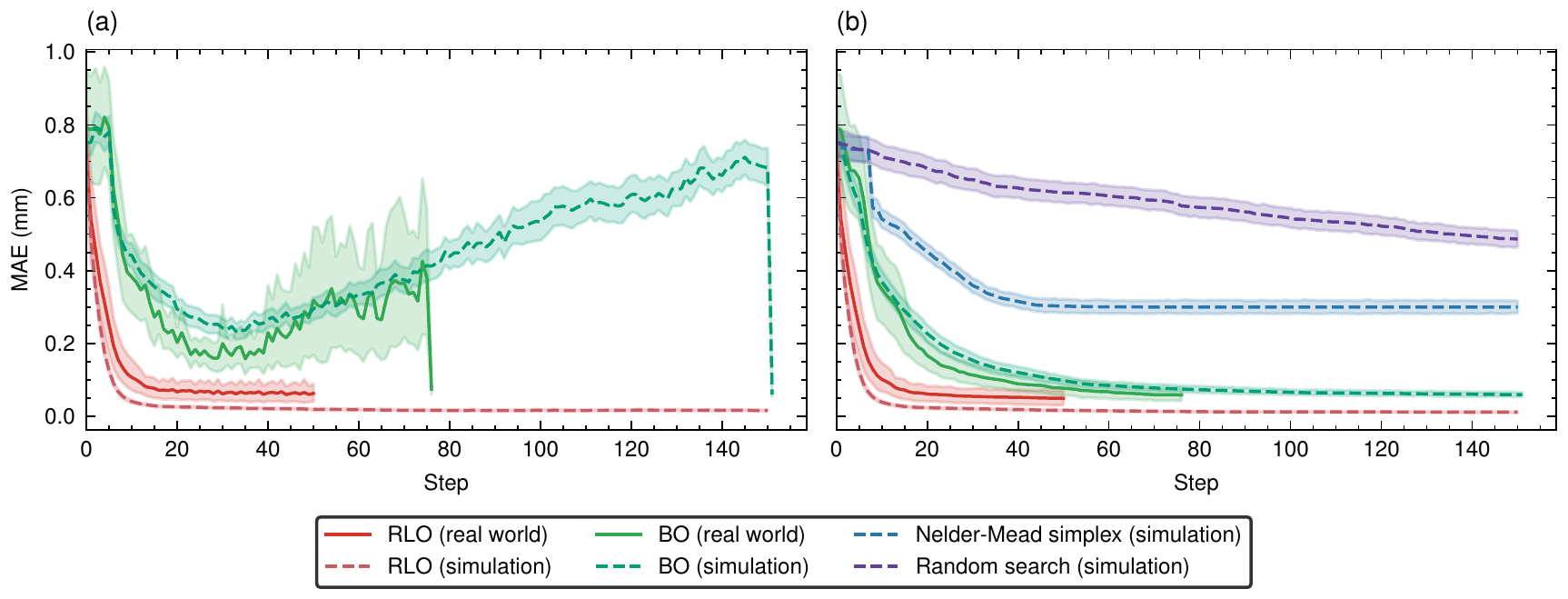}
    \caption{\textbf{Beam difference over time for different optimisation algorithms.} The mean beam difference as the \ac{MAE} of the beam parameters to the target beam is shown by the solid and dashed lines. The envelopes show the \qty{95}{\percent} confidence intervals of the beam differences. \textbf{a} shows the beam differences as measured at each step. \textbf{b} shows the best beam differences encountered up to each step, i.e. the beam differences that one would return to if the optimisation was terminated in the respective step. Note that on the real plant, this is an estimate, as the beam difference may not be exactly the same for the same set of actuator settings at different times.}
    \label{fig:mae_over_time_combined}
\end{figure}

Based on $\measurementaccuracy$, we construct two metrics to measure the optimisation speed. We define \textit{steps to target} as the number of steps until the observed beam parameters differ less than an average of $\measurementaccuracy$ from the target beam parameters, and \textit{steps to convergence} as the number of steps after which the average of the beam parameters never changes by more than $\measurementaccuracy$.
We observe that \ac{RLO} always converges and manages to do so close to the target in \qty{88}{\percent} of trials. \Ac{BO} also converges on almost all trials, but only does so close to the target in \qty{12}{\percent} of trials, taking about \num{4} times longer to do so.
\Cref{fig:mae_over_time_combined} indicates why: \Ac{BO} explores the optimisation space instead of fully converging toward the target beam. It is possible to suppress this behaviour 
by using an acquisition function that favours exploitation, 
but our experiments have shown that such acquisition functions do not perform well with noisy objective functions. If a sample of the objective value was too high as a result of noise, the surrogate model is likely to overestimate the objective near that sample, causing \ac{BO} to get stuck instead of finding the true optimum.
We further observe that \ac{RLO} converges more smoothly than \ac{BO}. While this has little effect in simulation, in the real world, smooth convergence has various advantages like limiting wear on the actuators. In the particle accelerator benchmark, smoother convergence limits the effects of magnet hysteresis, an effect where the ferromagnetic core of an electromagnet retains some magnetisation when the current in its coils is removed, reduced, or reversed. As a result of such effects, the objective function may become noisy or even shift, which is why avoiding them through smooth actuator changes generally stabilises the objective function and improves reproducibility.

\subsection{Real-world study}\label{sec:realworld_study}

In order to evaluate the methods' ability to transfer to the real world and to verify the results obtained in simulation, we also studied the performance of \ac{RLO} and \ac{BO} on the ARES particle accelerator.
This part of the study is crucial, as even with accurate simulations, the gap between simulation and the real world is often wide enough that algorithms performing well in simulation cannot be transferred to the real plant~\cite{dulac2019challenges}. We observed this gap between simulation and experiment in the early stages of training the \ac{RL} policy, where trained policies performed well in simulation but failed altogether in the real world. Similarly, when implementing \ac{BO} for the tuning task, implementations tuned for exploitation showed faster and better optimisation in simulation but failed during the experiment under real-world conditions.

\begin{figure}
    \centering
    \includegraphics{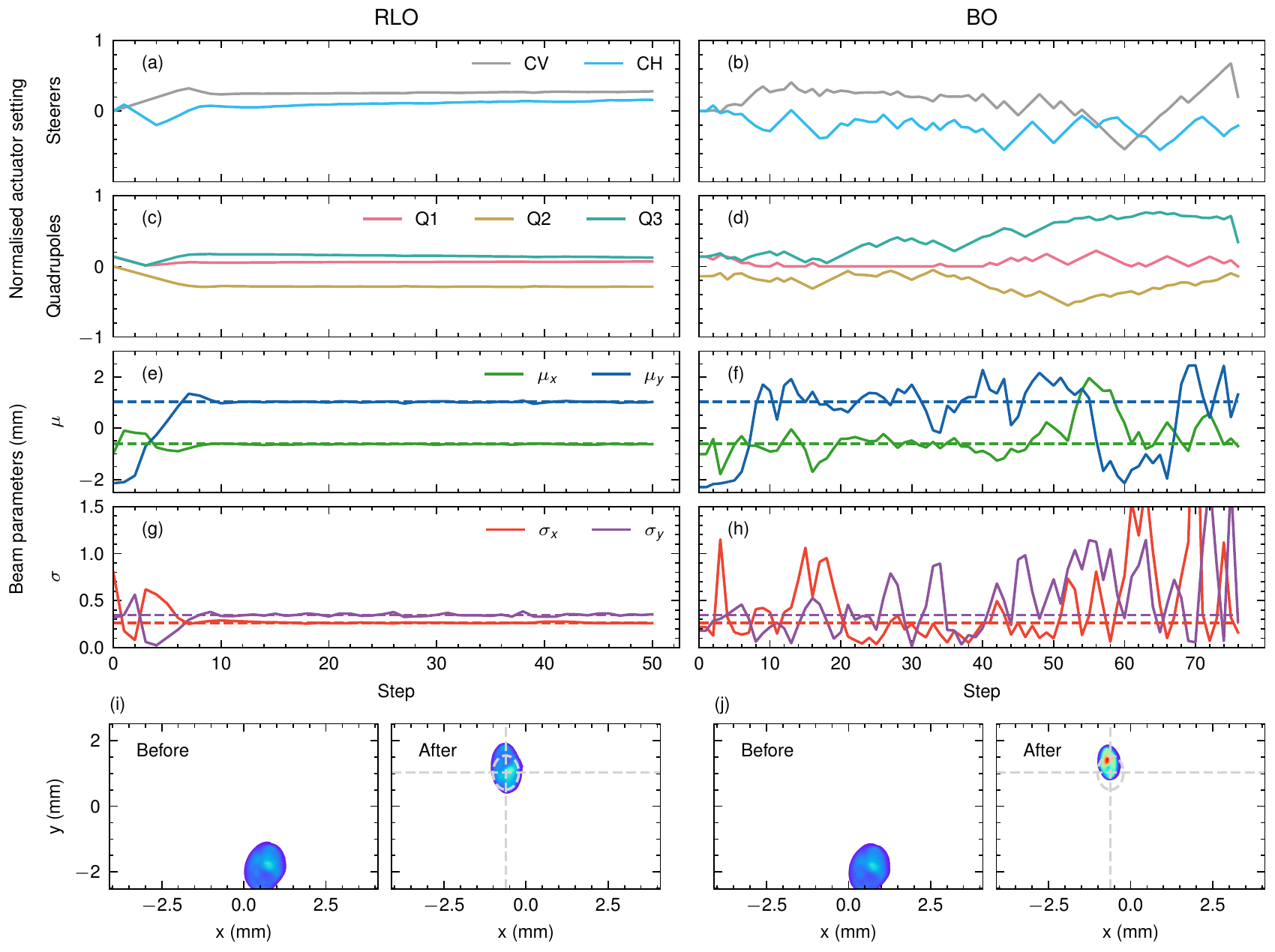}
    \caption{\textbf{Example optimisations on the real particle accelerator.} 
    \textbf{a,c,e,g} show one optimisation with \ac{RLO}. \textbf{b,d,f,h} show one optimisation with \ac{BO}.
    \textbf{a,b} show the steerer settings and \textbf{c,d} show the quadrupole magnet settings. \textbf{e,f} show the beam positions and \textbf{g,h} show the beam sizes.
    \textbf{i,j} show the beam images before and after the optimisation respectively. The target beam size and position are indicated with dashed lines.
    }
    \label{fig:example_episodes_combined}
\end{figure}

Given the limited availability of the real accelerator, we considered \nrealtrials trials of the \num{300} used for the simulation study. 
The magnet misalignments and the incoming beam on the real accelerator can neither be influenced nor measured during the experiment, so they were considered unknown variables. Before every measurement shift, the incoming beam was aligned with the centres of the quadrupole magnets in order to reduce dipole moments induced when the beam passes through a quadrupole magnet off-centre, which can steer the beam too far off the screen. This adjustment is needed for \ac{BO} to find an objective signal in a reasonable time. In \cref{sec:recovery_from_bad_inputs}, we investigate how the alignment, or the lack thereof, affects the results of this study. The results of the real-world measurements are listed in \cref{tab:results}. Two example optimisations by \ac{RLO} and \ac{BO} on the real accelerator are shown in \cref{fig:example_episodes_combined}. 
On the real particle accelerator, just like in the simulation study, we observe that \ac{RLO} achieves both a better tuning result and faster convergence than \ac{BO}. 
This time, \ac{RLO} outperforms \ac{BO} on \num{13} of \nrealtrials trials. The gap between the two, however, is not as pronounced in the real world. While all three performance metrics of \ac{BO} are almost identical between the real world and simulation, the performance of \ac{RLO} appears to degrade. 
This is partially due to the measurement accuracy now limiting the achievable beam difference, with the result of \ac{RLO} being only slightly larger than $\measurementaccuracy$ at \qty{24.46}{\micro\meter}. 
The degradation of \ac{RLO} performance may, however,  also be an indication that despite the use of domain randomisation the \ac{RL} policy has slightly overfitted on the simulation.
\Ac{BO} does not suffer from this issue as it learns at application time. Note also that to use the available machine study time most effectively, both algorithms were given fewer steps on the real accelerator than in simulation, and that \ac{BO} was given more steps than \ac{RLO}.

\subsection{Sim2real transfer}

The transfer of a method, that works well in a simulation environment, to the real-world is a large part of developing tuning algorithms for facilities such as particle accelerators. The challenges posed by this so-called \textit{sim2real} transfer impact the choice of tuning algorithm.

Successfully transferring a policy trained for \ac{RLO} to the real ARES accelerator involved a number of engineering decisions detailed in previous work~\cite{kaiser2022learningbased} and in~\cref{sec:reinforcement_learning}. While some of the design choices, such as inferring changes to the actuator settings instead of the actuator settings directly, can be applied to other tuning tasks with relative ease, others, such as domain randomisation~\cite{tobin2017domain,openai2019solving}, require specialised engineering for each considered tuning task. Furthermore, all of these require time-consuming fine-tuning to actually achieve a successful zero-shot sim2real transfer of a policy trained only in simulation. This is illustrated by the fact that many of the policies trained before the one studied here, performed excellently in simulation while sometimes not working at all on the real ARES accelerator.

On the other hand, \ac{BO} transfers to the real world with relatively little effort. Once it was sorted out how to best deal with faulty measurements, further discussed in~\cref{sec:recovery_from_bad_inputs}, most iterations of the \ac{BO} implementation performed about as well on the real accelerator as they did in simulation. Only some more specialised design decisions, such as tuning the acquisition function strongly towards exploitation, did not transfer as well. The easier sim2real transfer of \ac{BO} is likely owing to the fact that \ac{GP} model is learned entirely on the real plant and therefore will not overfit to a different objective function that deviates from the one under optimisation.

One issue that may arise when transferring \ac{BO} or random search from simulation to the real plant is that, while \ac{RLO} naturally converges toward an optimum and then stays there, meaning that if the optimisation is ended at any time the environment's state is at, or at least close to, the best-seen position, algorithms like \ac{BO} and random search are likely to explore further after finding the optimum. It is therefore necessary to return to the best-seen input when the optimisation is terminated. 
In simulation, this strategy will recover the same objective value, but real-world objective functions are noisy and not always perfectly stationary, e.g. due to slow thermal drifts.
As a result, effects such as magnet hysteresis on particle accelerators may shift the objective value when returning to a previously seen point in the optimisation space. In the benchmark tuning task, we experience noisy measurements and magnet hysteresis. We found that for the studied \ac{BO} trials, the final beam error deviated by a median of \qty{11}{\micro\meter} and a maximum of \qty{42}{\micro\meter}. This means that the deviation is usually smaller than the measurement accuracy $\measurementaccuracy$. At least for the benchmark task, the effect is therefore non-negligible, but also not detrimental to the performance of the tuning algorithms.

\subsection{Inference times}

The time it takes to infer the next set of actuator settings may also influence the algorithm choice. For the benchmark task, the inference time happens to be negligible, because our benchmarked physical system, specifically the magnets and the beam measurement, is orders of magnitude slower than the inference time. At other facilities, where the physical process takes less time, the time taken for tuning may be dominated by the inference time of the tuning algorithm and there might even be real-time requirements~\cite{wang2021accelerated}.

We measure the average inference times of both algorithms over the \num{45000} inferences of the simulation study using a MacBook Pro with an M1 Pro chip running Python 3.9.15. We observe that \ac{BO} takes an average of \qty{0.7}{\second} to infer the next actuator settings, while \ac{RLO} is more than three orders of magnitude faster at \qty{0.0002}{\second}. This is because the \ac{RLO} policy requires only one forward pass of the \ac{MLP} with a complexity of $O(1)$ with respect to the steps taken. By contrast, in each \ac{BO} inference step, a full optimisation of the acquisition function is performed. This involves inferences with the GP model with complexity $O(n^3)$, scaling with the number of steps taken $n$.
Note that the \ac{RLO} inference can be significantly sped up by using specialised hardware~\cite{scomparin2022kingfisher}.

\subsection{Robustness in the presence of sensor blind spots}\label{sec:recovery_from_bad_inputs}

In any real system, it is possible to encounter states 
where the available diagnostics deliver false or inaccurate readings, causing erroneous objective values and observations.
Transitions to these states can be caused by external factors as well as the tuning algorithm itself. A good tuning algorithm should therefore be able to recover from these states.
In the benchmark tuning task, an erroneous measurement occurs when the electron beam is not visible on the diagnostic screen within the camera's field of view. In this case, the beam parameters computed from the diagnostic screen image are false, also resulting in a faulty objective value.

We observed that when the beam is not properly observed, \ac{RLO} can usually recover it in just a few steps. Presumably, the policy can leverage its experience from training to make an educated guess on the beam's position based on the magnet settings even though faulty beam measurements were not part of its training, where \ac{RLO} always had access to the correct beam parameters.

In contrast, \ac{BO} struggles to recover the beam when it is off-screen, as the \ac{GP} model is learned at application time from faulty observations, resulting in faulty predictions of the objective and acquisition functions. When defining the task's objective function as only a difference measure from the current to the target beam, falsely good objective values are predicted in the blind spot region of the actuator space and \ac{BO} converges towards their locations. Our implementation, as described in~\cref{sec:bayesian_optimisation}, alleviates this issue by introducing a constant punishment to the objective function when no beam is detected in the camera's field of view. Nevertheless, the lack of information about the objective function's topology results in \ac{BO} taking many arbitrary steps before the beam is by chance detected and the optimisation starts progressing towards the target beam. While more comprehensive diagnostics can help solve this problem, these are often not available.

Because of \ac{BO}'s insufficient ability to recover from a system state in which there is no informative objective signal, the presented measurements on the real accelerator were taken with the beam aligned to the quadrupole magnets. As a result, the additional dipole moments induced by the quadrupole magnets when increasing the magnitude of their focusing strength are kept minimal, reducing the chance that the beam leaves the camera's field of view during the initial step of the optimisation.
As this alignment would not be performed during nominal operation but may change the observed performance of both algorithms, a study was performed in simulation in order to understand how to interpret the reported results given 
that the beam was aligned to the centres of the quadrupole magnets before the optimisation.
Both algorithms are evaluated over the same \num{300} trials as in~\cref{sec:simulation_study}. Unlike in the original simulation study, we also simulate erroneous beam parameter measurements when the beam position is detected outside the camera's field of view. Both algorithms are tested once with the original incoming beam and once with an incoming beam that was previously aligned to the quadrupole magnets.
The results are reported in~\cref{tab:results}.
We conclude that the reported results on the real particle accelerator would be expected to worsen by about \qtyrange{5}{33}{\percent} for \ac{RLO} and by \qtyrange{12}{121}{\percent} for \ac{BO} 
if the electron beam had not been aligned to the centres of the quadrupole magnets 
at the beginning of measurement shifts. This does not change how both algorithms compare to each other.

\subsection{Failure modes}

With tuning algorithms that are intended to be deployed without supervision to enable the autonomous operation of complex plants, it is important to understand how they might fail. We observe that over the entirety of this study, neither \ac{RLO} nor \ac{BO} ever produced a final beam that was worse than the beam before the optimisation.
Instead, both algorithms clearly improve the beam in most trials, with only a few trials being outliers where the objective was only slightly improved.
It was not possible to identify for either \ac{RLO} or \ac{BO} a cause for these outliers. Most likely, they are stochastic in nature, owing to the stochastic components of either algorithm.
That is, the \ac{RLO} policy presumably did not gain enough experience in some regions of the state space because they were not explored as much during training.
Similarly, \ac{BO} may be at a disadvantage when the randomly chosen initial samples are unfavourable.
We performed grid scans over target beams for both algorithms in simulation to confirm this through the presence of outliers in random locations of the target beam space. They further show that both algorithms perform worse when tasked with tuning towards large than when tasked with tuning towards a small beam, though this effect is subtle compared to the outliers.
The root cause of this observation is presumably the initialisation of the magnets in an \ac{FDF} pattern at the beginning of each optimisation with both \ac{RLO} and \ac{BO}. While creating a performance deficit for certain beams compared to others, this initialisation improves the overall performance of both algorithms.

There are two further failure modes that should be discussed for both algorithms. \Ac{RLO} can in rare cases enter an unstable state, in which the policy outputs oscillating actuator settings. These result in the beam parameters oscillating around the target. The cause of these oscillations is yet unknown.
We note that the oscillations are also produced by policies trained with different random seeds. \Ac{BO} may fail seriously when the beam is far away instead of just slightly off the screen before the start of the optimisation. In such a case, it can take a long time before the beam is randomly moved into the visible area of the diagnostic screen.

\subsection{Running as a feedback}

Real-world plants may be subject to drifts caused by unmodelled external factors. Moreover, control can be regarded as the continuous optimisation of a dynamic objective function.
Consequently, a tuning algorithm that can run as feedback on a dynamic objective function can be used for drift compensation and control in addition to tuning. Thus, a tuning method's ability to operate as a feedback is an interesting subject of further investigation and could impact algorithm selection.

\begin{figure}
    \centering
    \includegraphics{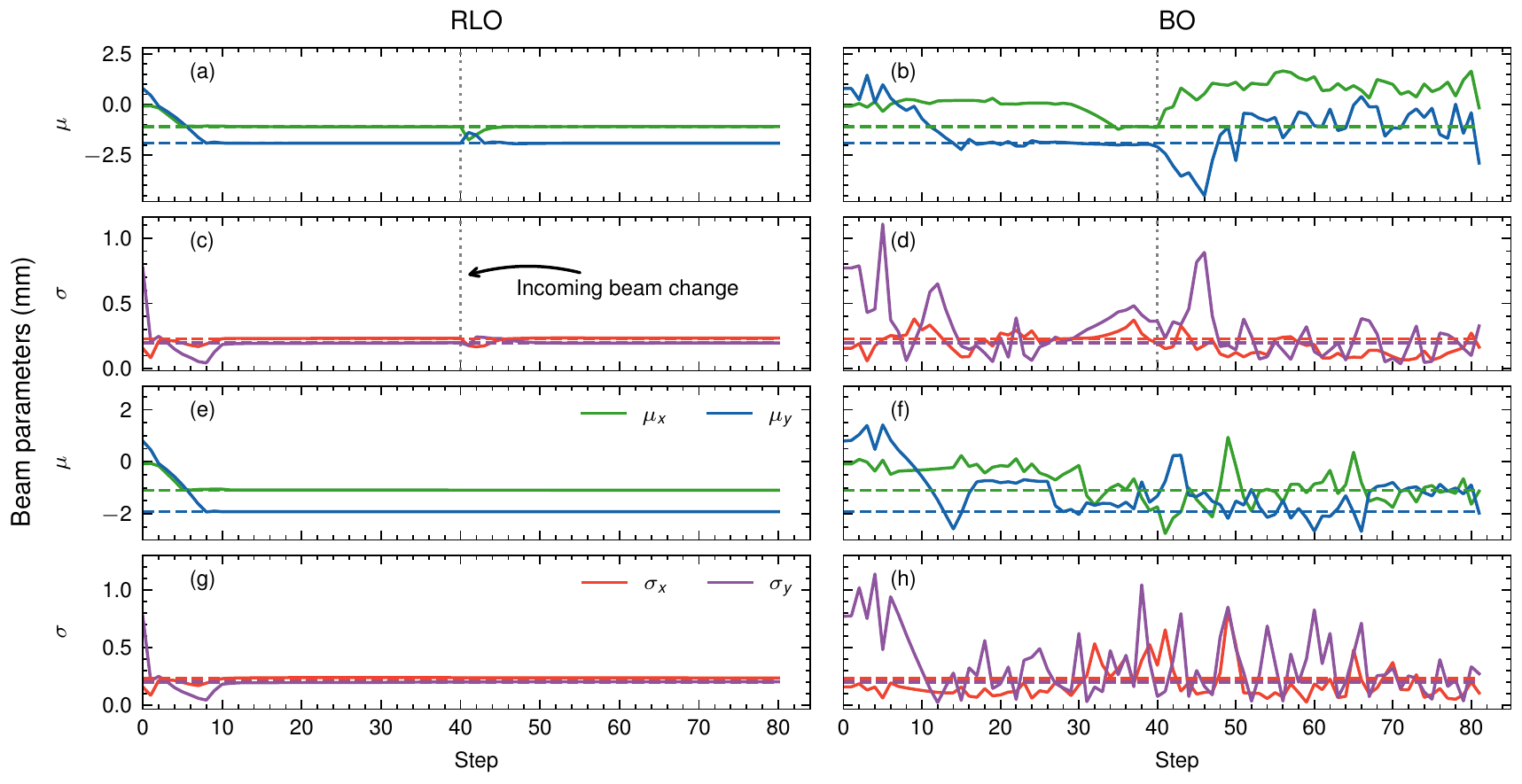}
    \caption{\textbf{\Ac{RLO} and \ac{BO} optimisers running as feedbacks in simulation.} 
    \textbf{a}-\textbf{d}, \ac{RLO} and \ac{BO} reacting to an instant change of the incoming beam at step \num{40}, denoted by the vertical dotted lines.
    \textbf{e}-\textbf{h}, the optimisers tracking the optimum with respect to a continuously changing incoming beam.
    \textbf{a,b,e,f} show the evolution of the beam positions $\{\mu_x,\mu_y \}$, and \textbf{c,d,g,h} show the beam sizes $\{\sigma_x,\sigma_y \}$. The horizontal dashed lines denote the target beam parameters respectively.
    }
    \label{fig:feedback_example_episodes_beam_parameters}
\end{figure}

While \ac{BO} assumes a static objective function, the benchmarked \ac{RLO} policy does not rely on memorising previously seen objective values or could alternatively learn to adapt to dynamic objective functions during training. It should therefore be possible to use the policy from \ac{RLO} as an \ac{RL}-based feedback controller for a dynamic system. 
To test this, we ran an optimisation with both methods for \num{80} steps. After \num{40} steps, we introduce an instant step-to-step change to the incoming beam, changing the latter such that a different set of actuator settings is required to achieve the same beam on the diagnostic screen.
We then observe how the \ac{RL} policy and our \ac{BO} implementation react to the upstream change. If the method manages to recover the machine state, 
it can be considered capable of running as feedback. 
As can be seen in \cref{tab:feedback_and_failure_results} and \cref{fig:feedback_example_episodes_beam_parameters}, the \ac{RL}-based controller can in fact recover the target beam in about the same time it took to perform the original optimisation, with the final beam difference being comparable to the one achieved in optimisation with a static incoming beam.
The beam achieved by \ac{BO} when the beam instantly changes during the optimisation is, as expected, significantly worse than it is with a constant beam. After the incoming beam changed, the \ac{GP} model based on the previous \num{40} samples is no longer correct, effectively breaking \ac{BO}.

However, the system changes that feedbacks need to react to are not always fast. Often, they occur slowly over time, such that the controller must track the change in order to hold the system near the desired state after attaining it. We therefore also evaluate the \ac{RLO} policy as a controller and \ac{BO} in a setup where the incoming beam changes linearly over the course of \num{80} steps. The results are listed in \cref{tab:feedback_and_failure_results}. We can see in \cref{fig:feedback_example_episodes_beam_parameters} that the \ac{RL} policy is capable of tracking the target beam parameters after attaining them.
The reasonably small increase in final \ac{MAE} can primarily be explained by the fact that the policy requires a few steps to converge on the desired beam parameters, but in the final step only a single step has passed since the last change to the incoming beam, therefore giving the policy only very little time to correct for the change. 
As with the instant incoming beam change, \ac{BO} is not capable of tracking the desired beam parameters. As the incoming beam cannot be included in the GP-model, the learned surrogate is ill-defined, tracking a dynamically changing objective function with a static model. As a result \ac{BO} optimises an objective function that diverges from the true objective function of the system.

It needs to be mentioned that the slow drifts of the underlying objective function, like the temperature drift of the magnets, can be tackled by adaptive BO with a spatiotemporal GP model~\cite{nyikosa2018adaptivebo} or contextual \ac{BO}~\cite{krause2011contextual}.
This would require, however, problem-specific implementation and additional engineering effort.

\subsection{Robustness to actuator failure}

In real-world plants, one also has to deal with the potential failure of components such as the actuators used for tuning. It would therefore be beneficial if a tuning algorithm could handle such an actuator failure and recovers the previous state.

\begin{figure}
    \centering
    \includegraphics{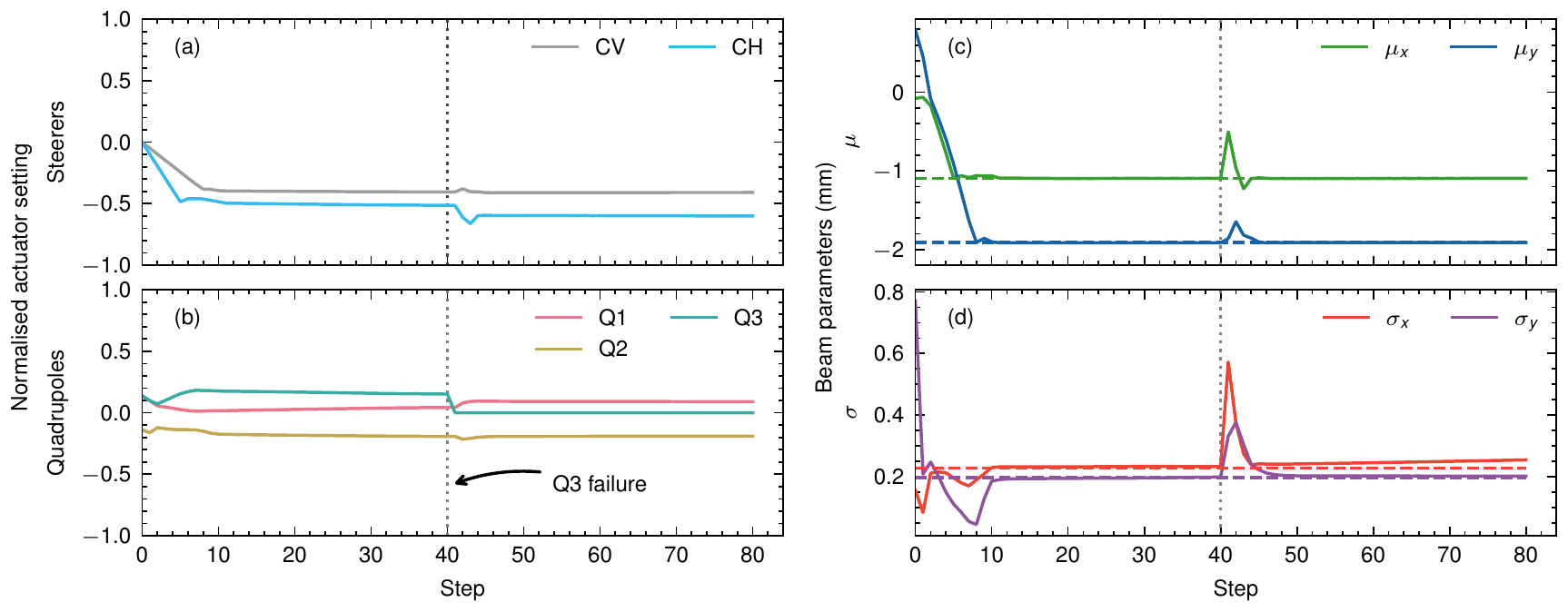}
    \caption{\textbf{\Ac{RLO} reacting to an simulated actuator failure during the optimisation.} The third quadrupole magnet fails in step \num{40}, denoted by the vertical dotted lines.
    \textbf{a} shows the normalised steerer settings and \textbf{b} showed the normalised quadrupole strengths, where, when Q3 fails, the strength of the other focusing quadrupole magnet Q2 is quickly increased and the horizontal steering magnet is used to counter the increased change beam position as a result of the changing dipole moments induced by the quadrupole magnets. \textbf{c} shows the beam positions $\{\mu_x, \mu_y \}$ and \textbf{d} shows the beam sizes $\{ \sigma_x, \sigma_y \}$. The dashed lines denote the target beam parameters respectively.}
    \label{fig:magnet_failure_rl_example}
\end{figure}

We evaluate \ac{RLO}'s and \ac{BO}' ability to handle both a permanent actuator failure, where the actuator has failed some time before the tuning algorithm was started, and a delayed actuator failure, where the actuator is operational when the tuning starts but fails at a later time during the tuning. Specifically, we simulate the failure of the third quadrupole magnet in the benchmark task, assuming that the magnet's power supply has failed and its quadrupole strength is permanently set to \qty{0}{\per\meter\squared} after the initial failure. We assume that a failed actuator provides a correct readback to the optimisation algorithm. \Cref{tab:feedback_and_failure_results} lists the results. We observe that \ac{RLO} handles actuator failure well, despite never being trained to do so. 
When the magnet has failed before the start of the optimisation, \ac{RLO} finds an optimum, that is almost as good as it would be without the magnet failure, without using the failed magnet. \Ac{RLO} can recover the beam's state when the magnet fails during the optimisation. An example of \ac{RLO} reacting to an actuator failure during tuning is shown in \cref{fig:magnet_failure_rl_example}. \Ac{BO} even improves in performance, as a failed magnet reduces the dimensions of the search space, but despite this, it performs worse than \ac{RLO}.

\section{Discussion}\label{sec:discussion}

\begin{figure}
    \centering
    \includegraphics{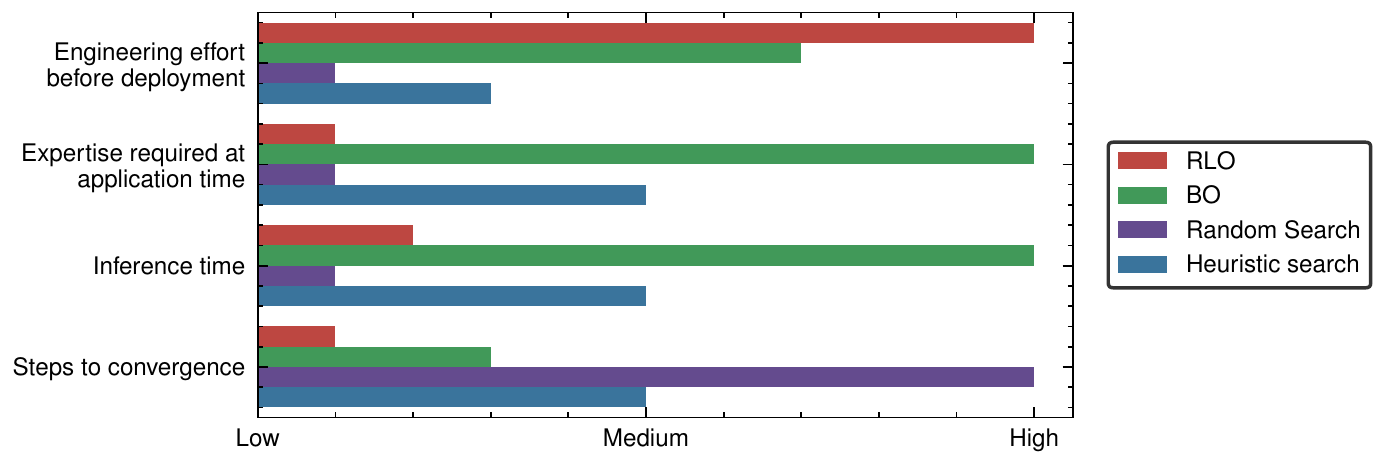}
    \caption{\textbf{Design space for large-scale facility tuning algorithms.} Shows qualitative metrics of comparison for all algorithms considered in our study relative to each other and may aid the decision-making process for choosing one of these algorithms based on criteria specific to the desired application.}
    \label{fig:design_space}
\end{figure}

The results of our study show that both learning-based optimisation algorithms \ac{RLO} and \ac{BO} clearly outperform the baseline methods Nelder-Mead Simplex optimisation and random search. Furthermore, the results indicate that in most cases, \ac{RLO} is the superior learning-based optimisation method, thanks to its ability to utilise experience acquired before the application time. Nevertheless, \ac{BO} proves to be a promising alternative for online continuous tuning of complex real-world plants due to its versatility as a black-box optimisation method.
In \cref{fig:design_space} we illustrate, how both learning-based algorithms relate to each other and the two investigated baselines in terms of different design aspects.

\Ac{RLO} primarily outperforms \ac{BO} in that it is capable of converging towards the desired plant state faster than \ac{BO} and closer to the desired state. Furthermore, \ac{RLO} was found to be more capable of dealing with many of the challenges encountered when working with real-world plants. 
When presented with false sensor readings, \ac{RLO} recovers faster than \ac{BO}. Furthermore, \ac{RLO} does not continue exploring once the optimum is found, eliminating the problems associated with recovering previously seen objective values in real-world systems. 
In addition, a trained \ac{RLO} agent requires no setting of hyperparameters or similar at application time and can therefore be used as a one-click solution by anyone without requiring \ac{RL} expertise from the user and promising reproducible results. This is in contrast to \ac{BO}, which is likely to need small hyperparameter adjustments for different instances of the same tuning tasks, thus requiring that a user brings at least some understanding of the chosen \ac{BO} implementation and its hyperparameters. While not the main focus of this study, policies trained via \ac{RL} as optimisers may also be used without retraining as controllers, being able to both reach the optimum in a static system and track the optimum in a dynamic system.

The main advantage of \ac{BO} is the relatively small engineering effort required to deploy it successfully. \ac{BO} algorithms that adapt hyperparameters automatically during the actual optimisation can be implemented easily and require relatively little hyperparameter tuning between different tuning tasks. In contrast, \ac{RLO} requires substantial engineering efforts by both \ac{RL} and domain experts, who must develop a suitable training setup and overcome the sim2real transfer problem. In addition, we observed that both \ac{RLO} and \ac{BO} can deal with unexpected situations like actuator failures, thus being robust tuning methods for real-world applications.

The choice of tuning algorithm depends primarily on how much and how often a tuning algorithm is going to be used, and whether the final tuning result and the time saved by a fast tuning algorithm are worth the associated engineering effort. We find that \ac{RLO} is the overall more capable and faster optimiser, but requires significant upfront engineering. It is therefore better suited to regularly performed tasks, where better tuning results and faster tuning justify the initial investment.
For tasks that are only performed a few times, for example on rare occasions during operation or during the commissioning of a system, the engineering effort associated with \ac{RLO} may not be justified. Our study has shown that \ac{BO}, despite not performing as well as \ac{RLO}, is a valid alternative in such cases.

\section{Methods}\label{sec:methods}

To collect the data presented in this study, evaluation runs of \ac{RLO} and \ac{BO} as well as the baseline methods of Nelder-Mead Simplex and random search were run in simulation and on a real particle accelerator. The following sections introduce the real-world plant used for our study, our experimental setups, and the optimisation algorithms.

\subsection{ARES particle accelerator section}\label{sec:ares_experimental_area}

The ARES (Accelerator Research Experiment at SINBAD) particle accelerator~\cite{burkart2022the,panofski2021commissioning}, located at Deutsches Elektronen-Synchrotron DESY in Hamburg, Germany, is an S-band radio frequency linac that features a photoinjector and two independently driven travelling wave accelerating structures. These structures can operate at energies up to \qty{154}{\mega\eV}. The primary research focus of ARES is to produce and study sub-femtosecond electron bunches at relativistic energies. The ability to generate such short bunches is of great interest for applications such as radiation generation by free electron lasers.
ARES is also used for accelerator component research and development as well as medical applications.

\begin{figure}
    \centering
    \includegraphics{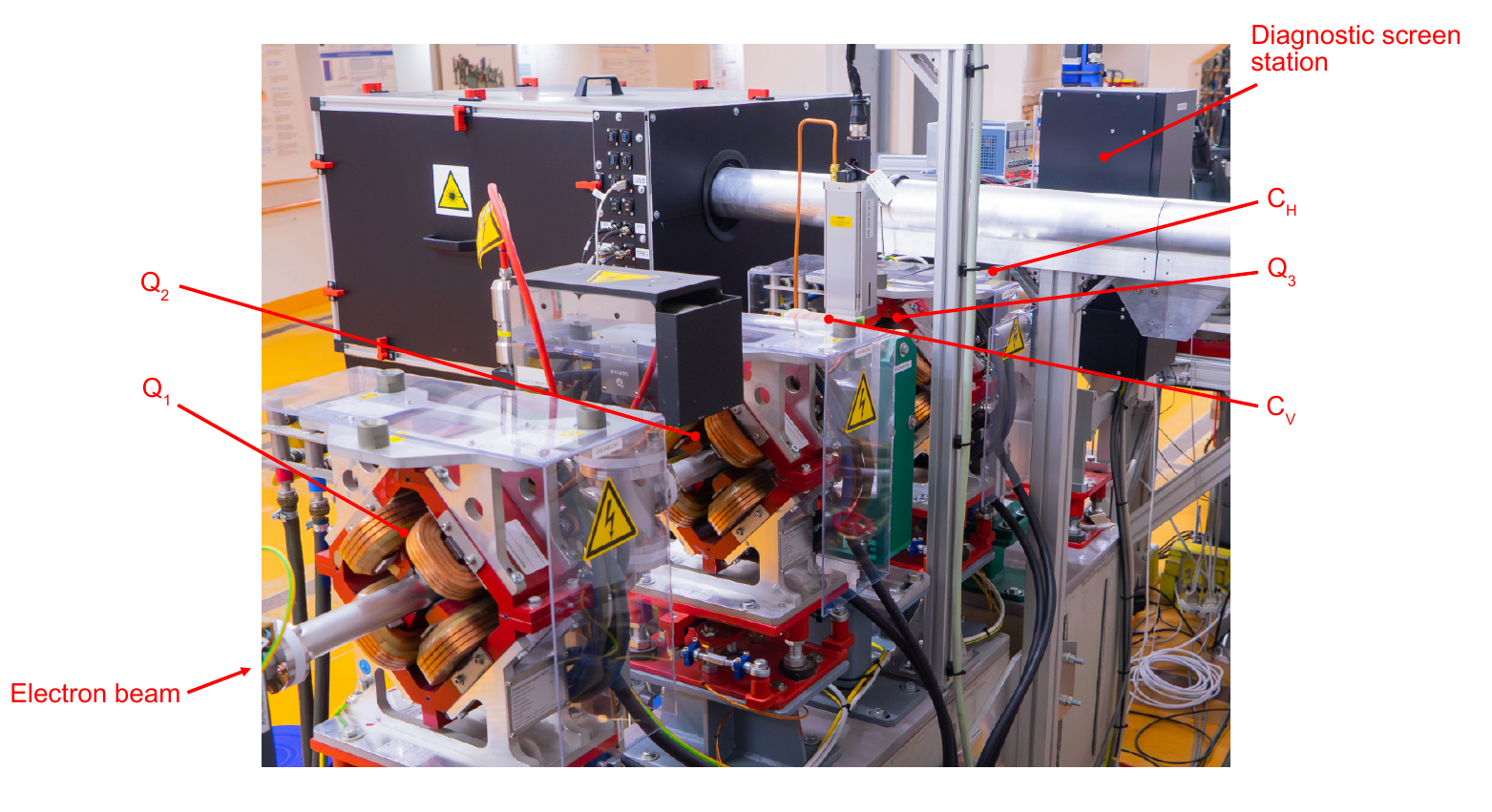}
    \caption{\textbf{Considered accelerator section at ARES.} The electron beam travels downstream from left to right. The five magnets actuated by the optimisation algorithms and the diagnostic screen station are marked.
    }
    \label{fig:ares_ea_cad}
\end{figure}

The accelerator section, known as the \textit{Experimental Area}, is a subsection of ARES, shown in \cref{fig:ares_ea_cad} and made up of two quadrupole magnets, followed by a vertical steering magnet that is followed by another quadrupole magnet and a horizontal steering magnet. Downstream of the five magnets, there is a scintillating diagnostic screen observed by a camera. The power supplies of all magnets can be switched in polarity. The quadrupole magnets can be actuated up to a field strength of \qty{72}{\per\meter\squared}. The limit of the steering magnets is \qty{6.2}{\milli\radian}. The camera observes an area of about \qty{8}{\milli\meter} by \qty{5}{\milli\meter} at a resolution of \num{2448} by \num{2040} pixels. The effective resolution of the scintillating screen is ca. \qty{20}{\micro\meter}.

At the downstream end of the section, there is an experimental chamber. This section is regularly used to tune the beam to the specifications required in the experimental chamber or further downstream in the ARES accelerator.

\subsection{Simulation evaluation setup}

In the simulation, a fixed set of \num{300} randomly generated trials were used to compare the different optimisation algorithms. Each trial is a tuple

\begin{equation}\label{eq:trial}
    \left( \targetbeam, M, I \right)
\end{equation}

of the target beam $\measuredbeam$ that we wish to observe on the diagnostic screen, the misalignments of the quadrupole magnets and the diagnostic screen $M$, as well as the incoming beam $I$ entering the accelerator section. The target beam was generated in a range of \qty{\pm 2}{\milli\meter} for $\mu_x$ and $\mu_y$, and \qtyrange{0}{2}{\milli\meter} for $\sigma_x$ and $\sigma_y$. These ranges were chosen to cover a wide range of measurable target beam parameters, which are constrained by the dimensions of the diagnostic screen. 
The incoming beam $\incomingbeam$ is randomly generated to represent samples from the actual operating range of the real-word accelerator. Both incoming beam and misalignment ranges were chosen to be larger than their estimated ranges present in the real machine.

\subsection{Real-world evaluation setup}

In the real world, the overall machine state was set to an arbitrary normal machine state, usually by leaving it as it was left from previous experiments. 
This should give a good spread over reasonable working points. The target beams were taken from the trial set used for the simulation study. 
As the incoming beam and misalignments cannot be influenced in the real world in the same way they can be in simulation, they are left as they are on the real accelerator and considered unknown. 
Experiments on the real accelerator were conducted on \num{9} different days over the course of \num{82} days, running at charges between \qty{2.6}{\pico\coulomb} and \qty{29.9}{\pico\coulomb}, and an energy of \qty{154}{\mega\eV}.
To ensure a fair comparison of the tuning methods, we align the beam to the quadrupole magnets at the beginning of each measurement day. This ensures that the beam remains within the camera's field of view on the diagnostic screen in the initial step, which is also a common operating condition of the accelerator. 
This reduces the dipole moments produced when increasing the strength of the quadrupole magnets and therefore reduces the likelihood of the beam being steered past the camera's field of view on the diagnostic screen in the very first step when \ac{BO} changes the quadrupole strengths. The alignment is not necessarily needed for the \ac{RLO} as it can recover the beam back into the diagnostic screen camera's field of view despite receiving erroneous observations.

\begin{figure}
    \centering
    \includegraphics[width=\textwidth]{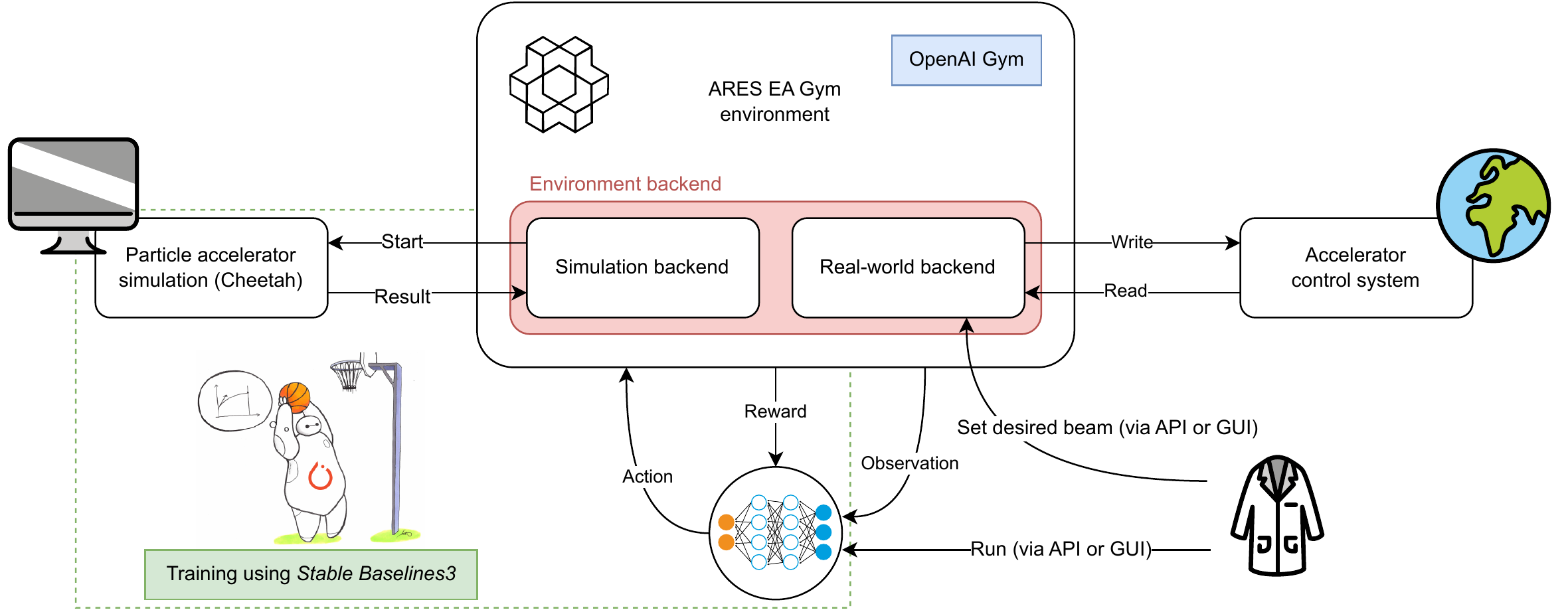}
    \caption{\textbf{Gym environment setup used for the study.} In particular showing how one environment interface facilitates design and training using a simulation of the plant, as well as transferring a developed tuning algorithm to the real plant without modification.}
    \label{fig:ares_ea_rl_technical_setup}
\end{figure}

Transferability of the experiments between simulation and real world as well as \ac{RLO} and black-box optimisation was achieved through a combination of OpenAI Gym~\cite{brockman2016openai} environments an overview of which is shown in \cref{fig:ares_ea_rl_technical_setup}. 
Two different environments were created based on a common parent environment defining the logic of the beam tuning task. One wraps around the \textit{Cheetah} simulation code~\cite{stein2022accelerating}, allowing for fast training and evaluation. The other environment interfaces with the accelerator's control system. Crucially, both environments present the same interface, meaning that any solution can easily be transferred between the two. 
While Gym environments are primarily designed for \ac{RL} policies to interact with their task, the ones used for this work were made configurable in such a way that they can also be interfaced with a \ac{BO} optimisation. This includes configurable reward formulations and action types that pass actuator settings to the \texttt{step} method and have the latter return an objective value via the \texttt{reward} field of the \texttt{step} method's return tuple.

\subsection{Reinforcement learning}\label{sec:reinforcement_learning}

The \ac{RLO} implementation used for this study has been introduced in previous work~\cite{kaiser2022learningbased}. In this case, an \ac{MLP} with two hidden layers of \num{64} nodes each is used as a policy, observing as input the currently measured beam parameters on the diagnostic screen $\measuredbeam = \left( \mu_x, \sigma_x, \mu_y, \sigma_y \right)$, the currently set field strengths and deflection angles of the magnets \mbox{$\actuators = \left(\quadstrength_{Q_1}, \quadstrength_{Q_2}, \kick_{C_v}, \quadstrength_{Q_3}, \kick_{C_h} \right)$} and the desired beam parameters $\bm{b}' = \left( \mu_x, \sigma_x, \mu_y, \sigma_y \right)$ set by the human operator. 
The policy then outputs changes to the magnet settings $\action_t = \Delta \bm{u}$.  
A normalisation of rewards and observations using a running mean and standard deviation is performed over the training. The outputs are normalised to 0.1 times the magnet ranges of \qty{\pm30}{\per\meter\squared} for the quadrupole magnets and \qty{\pm2}{\milli\radian} for the steering magnets. During training and application, optimisations are started from a fixed \ac{FDF} setting of the quadrupole triplet, with the strengths $(\quadstrength_{Q_1}, \quadstrength_{Q_2}, \quadstrength_{Q_3})=(10,-10,10)$ \unit{m^{-2}} and both steering magnets set to \qty{0}{\milli\radian}. The policy is trained for \num{6000000} steps using the \ac{TD3}~\cite{fujimoto2018addressing} algorithm as implemented by the \textit{Stable Baselines3}~\cite{raffin2019stable} package. Training is run in a simulation provided by the \textit{Cheetah}~\cite{stein2022accelerating} particle tracking code, as limited availability makes training on the real particle accelerator infeasible. Domain randomisation~\cite{tobin2017domain} is performed during training. Specifically, the magnet and screen misalignments, the incoming beam and the target beam are randomly sampled from a uniform distribution for each episode. The reward function used for training is

\begin{equation}
    R\left(\state_t, \action_t\right) = \begin{cases}
        \hat{R} \left( \state_t, \action_t \right) & \text{if $\hat{R} \left(\state_t, \action_t\right) > 0$} \\
        2 \cdot \hat{R} \left(\state_t, \action_t\right) & \text{otherwise}\,
    \end{cases}
\end{equation}

with $\hat{R} \left(\state_t, \action_t\right) = O\left(\actuators_t\right) - O\left(\actuators_{t+1}\right)$ and $O\left(\actuators_t\right)$ being the natural logarithm of the weighted \ac{MAE} between observed and target beam on the diagnostic screen. The trained policy is deployed zero-shot, i.e. without any further training or fine tuning, to the real world.

\subsection{Bayesian optimisation}\label{sec:bayesian_optimisation}

The \ac{BO} version used for this study is a custom implementation using the \textit{BoTorch}~\cite{balandat2020botorch} package. The objective $O(\actuators)$ to be optimised is defined as

\begin{equation}
    O(\actuators) = -\log \left( \text{MAE}(\bm{b}, \bm{b}') \right) + w_\text{on-screen}.
\end{equation}

The logarithm is used to properly weigh the fine improvement when \ac{BO} approaches the target beam. A further on-screen reward $w_\text{on-screen} = 10$ is added to the objective when the beam can be observed on the screen, and subtracted from the objective to penalise the settings when the beam is off the diagnostic screen.
To increase the numerical stability of the \ac{GP} regression, the previous input settings $\actuators$ are normalised to $[-1,1]$, projecting the maximum to \num{1} and the minimum to \num{-1}, and objective values are standardised. The covariance function of the GP models used in this study is the sum of a Matérn-5/2 kernel~\cite{matern1986spatial} and a white noise function. The GP hyperparameters, like the length scales and signal noise, are determined dynamically by log-likelihood fits in each step.
In each trial, \ac{BO} is started from the same fixed \ac{FDF} setting used by \ac{RLO}. Five random samples are taken to initialize the \ac{GP} model. 
Based on the posterior prediction of the GP model, an \ac{EI}~\cite{jones1998bayesian} acquisition function is calculated, which automatically balances the exploration and exploitation of the objective. The next sample is chosen by maximising the acquisition function, where the maximum step sizes are constrained to 0.1 times the total action space. Additionally, the quadrupole magnets are only allowed to vary unidirectionally, i.e. in the \ac{FDF} setting, so that the time-consuming polarity changes of the quadrupole magnets' power supplies due to the exploration behaviour of \ac{BO} can be avoided.
\ac{BO} is allowed to run \num{150} steps in simulation and \num{75} steps on the real machine, after which we return to the best settings found.

Note that this designed mostly using a simulation before deploying it to the real accelerator. This was done in an effort to reduce the amount of beam time needed for development.

\subsection{Nelder-Mead simplex}\label{sec:nelder_mead}

The Nelder-Mead Simplex optimisation~\cite{nelder1965simplex} was implemented using the \textit{SciPy}~\cite{virtanen2020scipy} Python package. The initial simplex was tuned in a random search of \num{405} samples to the one that performed best across the set of \num{300} trials. Nelder-Mead is allowed to run for a maximum of \num{150} steps. After \num{150} steps or after early termination the simplex might perform better if it returns to the final sample, but as it is generally converging, it does not necessarily need to. 
The objective function optimised by Nelder-Mead is the \ac{MAE} of the measured beam parameters to the target beam parameters.

\subsection{Random search}\label{sec:random_search}

For the random search baseline, we sample random magnet settings from the constrained space of magnet settings. Constraint in this case means that we limit the space to a range commonly used during operations, instead of the full physical limits of the magnets. The latter limits are almost an order of magnitude larger than anything ever used in operation. At the end of the optimisation, we return to the best example found.

\section*{Data Availability}

The data generated for the presented study is available at \url{https://doi.org/10.5281/zenodo.7853721}.

\section*{Code Availability}

The code used to conduct and evaluate the presented study is available at \url{https://github.com/desy-ml/rl-vs-bo}.

\section*{Acknowledgements}

This work has in part been funded by the IVF project InternLabs-0011 (HIR3X) and the Initiative and Networking Fund by the Helmholtz Association (Autonomous Accelerator, ZT-I-PF-5-6).
All figures and pictures by the authors are published under a CC-BY7 license. 
The authors thank Sonja Jaster-Merz and Max Kellermeier of the ARES team for their great support during shifts as well as always insightful brainstorms.
In addition, the authors acknowledge support from DESY (Hamburg, Germany) and KIT (Karlsruhe, Germany), members of the Helmholtz Association HGF, as well as support through the \textit{Maxwell} computational resources operated at DESY and the \textit{bwHPC} at SCC, KIT.

\section*{Author Contributions}

J.K., C.X., A.S.G., A.E. and E.B developed the concept of the study.
A.E., E.B and H.S. secured funding. 
J.K. developed and trained the \ac{RL} agent with support from O.S. 
C.X. designed the implementation of \ac{BO}. 
J.K. ran the simulated evaluation trials and took the real-world data. 
J.K. evaluated the measured data. 
C.X. provided substantial input to the evaluation. 
A.E. and A.S.G. provided input on the evaluation of the measured data. 
W.K., H.D., F.M. and T.V. assisted the data collection as ARES operators. 
F.B. assisted the data collection as ARES machine coordinator. 
W.K., H.D., F.M., T.V. and F.B. contributed their knowledge of the machine to the implementation of both methods. 
J.K. wrote the manuscript. 
C.X. provided substantial edits to the manuscript. 
J.K. created the presented figures with input from C.X., O.S. and F.M. 
All authors discussed the results and provided edits and feedback on the manuscript.

\section*{Competing Interests}

The authors declare no competing interests.

\nocite{*}

\bibliography{references} 

\clearpage

\begin{table}[h]
    \begin{threeparttable}
        \centering
        \caption{\textbf{Performance of optimisation algorithms on the benchmark tuning task.}}
        \label{tab:results}
        \begin{tabular}{lcccccccc}
            \toprule
            \textbf{Optimiser} & \multicolumn{2}{c}{\textbf{Final beam difference (\unit{\micro\meter})}} & \multicolumn{3}{c}{\textbf{Steps to target}} & \multicolumn{3}{c}{\textbf{Steps to convergence}} \\
            & Median & Mean & Median & Mean & Success rate & Median & Mean & Success rate \\
            \midrule
            \rowcolor{lightgray}
            \multicolumn{9}{c}{\textit{\textbf{Simulation} (infinite diagnostic screen, no beam alignment) \textrightarrow\ \cref{sec:simulation_study}}} \\
            \midrule
            Random search & \num{460} & $\num{490} \pm \num{200}$ & - & - & \qty{0.0}{\percent} & \num{33} & $\num{51} \pm \num{51}$  & \qty{100.0}{\percent} \\
            Nelder-Mead simplex & \num{270} & $\num{300} \pm \num{160}$ & \num{56} & $\num{56} \pm \num{0}$ & \qty{0.3}{\percent} & \num{29} & $\num{28} \pm \num{15}$ & \qty{100.0}{\percent} \\
            \ac{RLO} & \num{4} & $\num{11} \pm \num{19}$ & \num{9} & $\num{16} \pm \num{20}$ & \qty{88}{\percent} & \num{7} & $\num{9} \pm \num{11}$ & \qty{100.0}{\percent} \\
            \ac{BO} & \num{45} & $\num{60} \pm \num{55}$ & \num{52} & $\num{61} \pm \num{34}$ & \qty{12}{\percent} & \num{32} & $\num{42} \pm \num{31}$ & \qty{99.7}{\percent} \\
            \midrule
            \rowcolor{lightgray}
            \multicolumn{9}{c}{\textit{\textbf{Real world} (finite diagnostic screen, beam aligned to quadrupole magnets) \textrightarrow\ \cref{sec:realworld_study}}} \\
            \midrule
            \ac{RLO}\footnotemark[1]\footnotemark[2] & \num{24} & $\num{29} \pm \num{48}$ & \num{15} & $\num{17} \pm \num{11}$ & \qty{41}{\percent} & \num{10} & $\num{10.4} \pm \num{5.5}$ & \qty{100.0}{\percent} \\
            \ac{BO}\footnotemark[1]\footnotemark[3] & \num{44} & $\num{58} \pm \num{41}$ & \num{52.5} & $\num{52.5} \pm \num{5.5}$ & \qty{10}{\percent} & \num{33} & $\num{33} \pm \num{16}$ & \qty{100.0}{\percent} \\
            \midrule
            \rowcolor{lightgray}
            \multicolumn{9}{c}{\textit{\textbf{Simulation} (finite diagnostic screen, no beam alignment) \textrightarrow\ \cref{sec:recovery_from_bad_inputs}}} \\
            \midrule
            \ac{RLO} & \num{4} & $\num{13} \pm \num{40}$ & \num{8} & $\num{13} \pm \num{13}$ & \qty{85}{\percent} & \num{7} & $\num{8.2} \pm \num{6.8}$ & \qty{99.3}{\percent} \\
            \ac{BO} & \num{38} & $\num{61} \pm \num{96}$ & \num{45} & $\num{45} \pm \num{14}$ & \qty{10}{\percent} & \num{28} & $\num{32} \pm \num{15}$ & \qty{83.3}{\percent} \\
            \midrule
            \rowcolor{lightgray}
            \multicolumn{9}{c}{\textit{\textbf{Simulation} (finite diagnostic screen, beam aligned to quadrupole magnets) \textrightarrow\ \cref{sec:recovery_from_bad_inputs}}} \\
            \midrule
            \ac{RLO} & \num{4} & $\num{10} \pm \num{18}$ & \num{8} & $\num{13} \pm \num{13}$ & \qty{87}{\percent} & \num{7} & $\num{7.3} \pm \num{5.2}$ & \qty{99.7}{\percent} \\
            \ac{BO} & \num{23} & $\num{28} \pm \num{19}$ & \num{40} & $\num{40} \pm \num{13}$ & \qty{35}{\percent} & \num{25} & $\num{26} \pm \num{10}$ & \qty{97.0}{\percent} \\
            \bottomrule
        \end{tabular}
        \begin{tablenotes}
            \item \footnotesize The results reported in simulation have been collected over \num{300} different trials, for each of which the algorithms were given \num{150} steps for the optimisation. Because of limited beam time availability at the real accelerator, \num{22} trials\footnotemark[1] were selected for the real-world evaluations, \ac{RLO} was given \num{50} steps\footnotemark[2] for the optimisation and \ac{BO} was given \num{75} steps\footnotemark[3] for the optimisation. The final beam difference is computed as the \ac{MAE} over the beam parameters at the end of the optimisation according to \cref{eq:mae}. Steps to target are computed as the number of steps it took until an \ac{MAE} to the target beam smaller than the measurement accuracy of $\epsilon = \qty{20}{\micro\meter}$ was achieved. Steps to convergence are defined as the number of steps after which the improvement of the best seen \ac{MAE} remains smaller than the measurement accuracy~$\epsilon$. For each metric, we report the median and the mean with standard deviation. Both steps to target and steps to convergence metrics are reported only over trials for which the target or convergence, respectively, were achieved before the optimisation was terminated. We therefore also report success rates, that is the proportion of trials for which the target or convergence, respectively was successfully achieved before termination.
        \end{tablenotes}    
    \end{threeparttable}
\end{table}

\begin{table}[h]
    \begin{threeparttable}
        \centering
        \caption{\textbf{Results of feedback and actuator failure studies.}}
        \label{tab:feedback_and_failure_results}
        \begin{tabular}{lccccc}
            \toprule
            \textbf{Optimiser} & \textbf{Normal} & \multicolumn{2}{c}{\textbf{Feedback}} & \multicolumn{2}{c}{\textbf{Magnet failure}} \\
            & & Instant & Continuously & Before & During \\
            \midrule
            \ac{RLO} & \num{16} & \num{18} & \num{80} & \num{40} & \num{16} \\
            \ac{BO} & \num{73} & \num{350} & \num{250} & \num{45} & \num{50} \\
            \bottomrule
        \end{tabular}
        \begin{tablenotes}
            \item \footnotesize The values reported are mean final beam differences as \acp{MAE} from the observed beam parameters to the target beam parameters in \unit{\micro\meter}. All optimisations reported in this table were terminated at \num{80} steps. We therefore also report results for normal tuning runs where the upstream beam changes or magnet failures were introduced.
        \end{tablenotes}
    \end{threeparttable}
\end{table}

\end{document}